# Experimental Analysis Regarding the Influence of Iris Segmentation on the Recognition Rate

Heinz Hofbauer[1] • Fernando Alonso-Fernandez[2] • Josef Bigun[2] • Andreas Uhl[1]

[1]University of Salzburg, Department of Computer Sciences, e-mail: {hhofbaue, uhl}@cosy.sbg.ac.at
[2]Halmstad University, Intelligent Systems Lab, e-mail: {feralo, josef.bigun}@hh.se

## Abstract

In this paper we will look at the detection and segmentation of the iris and its influence on the overall performance of the iris-biometric tool chain. We will examine whether the segmentation accuracy, based on conformance with a ground truth, can serve as a predictor for the overall performance of the iris-biometric tool chain. That is: If the segmentation accuracy is improved will this always improve the overall performance? Furthermore, we will systematically evaluate the influence of segmentation parameters, pupillary and limbic boundary and normalization center (based on Daugman's rubbersheet model), on the rest of the iris-biometric tool chain. We will investigate if accurately finding these parameters is important and how consistency, i.e., extracting the same exact region of the iris during segmenting, influences the overall performance.

## Contents

# 1 Introduction

Iris recognition challenges for on-the-move and less constraint acquisitions, like the Noisy Iris Challenge Evaluation (NICE) [1], and Multiple Biometrics Grand Challenge (MBGC), illustrated the importance of robust iris segmentation in latest-generation iris biometric systems. Segmentation is a critical part in iris recognition systems (Fig. 1), since errors in this initial stage are propagated to subsequent processing stages.

Although state-of-the-art iris features are very effective for recognition, their performance is greatly affected by iris segmentation [2]. It is reported that most failures to match in iris recognition result from inaccurate segmentation [3]. Several factors can potentially degrade iris images [4, 5]. However, evaluation of its individual effect in the segmentation performance is quite limited [6, 7], with most of the works focused on its impact in the recognition accuracy [4, 8–10].

Early works on iris segmentation include the Daugman's approach using an integro-differential operator [11] and the method of Wildes involving the circular Hough transform [12]. They assume that iris boundaries can be approximated as circles. Much of the subsequent research has tried to improve the Wildes idea, such as the inherent computational burden of the Hough transform or the lack of enough edge points to define a circle [13]. Newer approaches for iris segmentation in less controlled environments relax the circularity assumption of circular boundaries, but many start with a detector of circular edges, which is further deformed into non-round boundaries. This is the case for example of active contours (also by proposed Daugman) [14], elastic models plus spline-based edge fitting [2] or AdaBoost eye detection [15]. Other approaches not relying initially on geometric models for detection, such as Graph Cuts [16], also make use of some circular of elliptical fitting during a refinement stage.

Iris recognition have been widely performed using sensors in the near-infrared (NIR) spectrum, since this type of lightning reveals the details of the iris texture better than sensors working with visible (VW) light [11]. However, the increasing prevalence of relaxed or uncooperative setups such as acquisition at-a-distance and on-the-move, where NIR illumination is unfeasible, has motivated the appearance of a body of research focused on the use of images captured with VW light [17]. A remarkable difference is that the inner (pupillary) iris boundary is usually more distinguishable than the outer (limbic) iris boundary with NIR images, whereas the opposite happens in VW light.

Research on iris segmentation constitutes a significant part of the published work in iris biometrics [18]. In addition to these, a variety of approaches are focused as well on finding perturbations that may occlude the iris region, such as specular reflections, eyelashes, or eyelids. For a coverage of existing techniques in these areas, we refer the reader to the surveys in [13, 18]. These occlusions, along with further artefacts like compression [19–21], also influence the performance of the iris biometric tool-chain.

Modern iris recognition algorithms operate on normalised representations of the iris texture ("Faberge coordinates" [11]), independent of pupillary dilation. For the mapping, parameterisations of inner and outer iris boundaries $P, L : [0, 2\pi) \to [0, m] \times [0, n]$ are employed to create a normalised iris texture via rubbersheet mapping $R(\theta, r) := (1 - r) \cdot P(\theta) + r \cdot L(\theta)$ using angular $\theta$ and pupil-to-sclera radial $r$ coordinates, such that a normalised texture and noise mask $T, M : [0, 2\pi) \times [0, 1] \to C$ are obtained ($C$ is the target color space, $M = N \circ R, T = I \circ R$ for the original $n \times m$ image $I$ and noise mask $N$). The latter usually considers reflections and upper and lower eyelid curves masking out occlusions, such that $N(x, y) \neq 0$ if and only if pixel $(x, y)$ refers to an in-iris location.

Since $P$ and $L$ are at the core of the iris biometric tool chain, the performance of iris segmentation algorithms is thought to be paramount to the performance of the overall system. The availability of ground truth data for iris segmentation [22] provides a tool to improve segmentation algorithms. However, it is unclear how exactly the segmentation influences the overall biometric system.

In this paper we will try to answer two questions: Can the segmentation performance, based on ground truth, predict the quality of the overall biometric system; And, what properties of the segmentation influence the overall biometric system?

In Section 2 we will describe the tools and datasets used for the evaluation purposes. Section 3 deals with the question whether or not the segmentation performance is a predictor for the performance of the overall iris biometric tool chain. Section 4 looks at the influence of the segmentation on the biometric tool chain by way of a controlled modification of the segmentation results. Finally, Section 5 concludes the paper.

# 2 Experimental Setup and Tools

To get a reliable picture of the behavior of the iris tool chain we opted to take as large a number of algorithms/datasets into account as is feasible. This is especially true for Section 3, however for Section 4 we choose to only use a subset of the following tools in order to reduce computational requirements since the experiments are rather extensive.

Evaluations will use ground-truth segmentation information from the IRISSEG-EP dataset [22, 23], specifically those of Operator A (*Op. A*). Note that only a subset of the *notredame* database has ground truth information, and consequently we only used this subset for experimentation. That said, we will use the following databases for our tests:

- **CASIA version 4.0 Iris-Interval subset** [24] (*casia4i*)
  A total of 2655 iris images are used from 249 subjects, both left and right eyes. The recording was done in the NIR spectrum in an indoor environment with a frontal view.
- **IIT Delhi Iris Database** [25] (*iitd*)
  A total of 2240 iris images are used from 224 subjects with 5 images per eye. The iris images are recorded in the near infrared (NIR) spectrum in an indoor environment with a frontal view.
- **Notredame ND-0405 Iris Image Dataset** [26] (*notredame*)
  A total of 837 images from 30 different subjects are used out of a total of 64980 images in the original database. The recording was done in the NIR spectrum in an indoor environment with a frontal view.

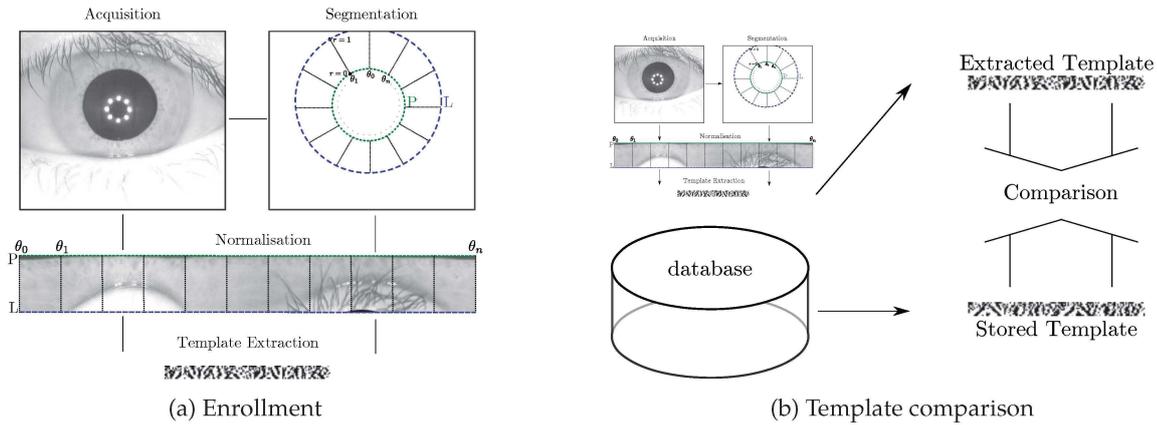

(a) Enrollment  (b) Template comparison

Figure 1: Biometric System overview

However, the Notredame database contains examples of 'real-world' type images with off-angle, blurry and off center recordings as well interlacing due to motion.

For the evaluation in the following sections we always utilized the databases to their full extent and included all comparisons in the evaluation which leads to the following number of genuine and imposter comparisons: *casia4i* with 8932 genuine and 3471909 imposter comparisons; *iitd* with 4480 genuine and 2503200 imposter comparisons; *notredame* with 11808 genuine and 338058 imposter comparisons.

For the various steps in the iris tool-chain, iris segmentation and feature extraction, we mainly relied on the open source implementations provided by the University of Salzburg Iris Toolkit (USIT) [27, 28], marked with ‡ in the following list. In addition we also supplemented the, partially older algorithms from the USIT, with algorithms from more recent literature. The tools used are:

1. Iris segmentation:
   - Contrast-adjusted Hough transform [27] (*CAHT*‡)
   - Weighted adaptive Hough and ellipsopolar transform [29] (*WAHET*‡)
   - The BioSecure reference system Osiris [30] (*Osiris*)
   - Iterative Fourier push-pull extraction [31] (*IFPP*‡)
   - Generalized structure tensor [32] (*GST*)

2. Feature extraction:
   - 1D Log-Gabor filters [33] (*lg*‡)
   - Quadratic spline wavelets [3] (*qsw*‡)
   - Cumulative sums of Gray-scale blocks [34] (*ko*‡ after the implementation name in USIT)
   - Local intensity variations within texture stripes [35] (*cr*‡ after the implementation in USIT)
   - Scale-invariant feature transform key points [36] (*sift*)
   - Discrete cosine transform based features [37] (*dct*‡)

The USIT specifically was designed to be used 'out of the box', as such there are basically no options to adjust to a specific dataset. Rather, the programs in the USIT try to estimate the best parameters based on the input without human interaction. We used the default parameters for all USIT software as well as for all other software used. There is only one exception: the verification tools do allow for a rotation compensation. Given that none of the databases are rotationally aligned the use of this option is a requirement, we set it to 7 bits left and right shift in the iris code, as suggested by [27] and the help screen of the tools (the parameter was -s -7 7).

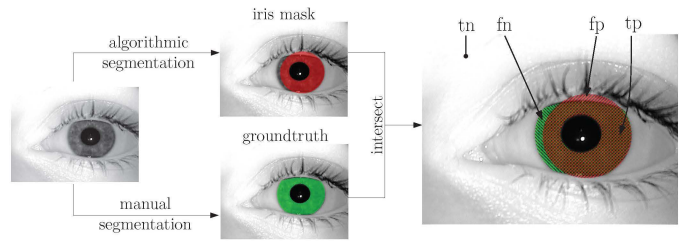

Figure 2: Example of the split between $tp$, $fp$, $tn$ and $fn$ for an iris image segmentation.

A note on other factors influencing the segmentation and overall performance like eyelids, eye lashes, reflections and obfuscation by hair and others. We will not include these factors in the systematic evaluation since a) their appearance in the source material can not be removed for a clean baseline and b) they can not easily be added artificially to facilitate a systematic approach to asses their influence.

We will also not use noise masks marking obfuscations of the iris since not all the tools used support the use and generation of noise masks. As such using them for some of the evaluations and not for other would make a direct comparison hard. We choose to not use the masks to be able to compare all tools equally.

There are numerous ways to calculate the segmentation error. The usual way to do this is to differentiate between the number of correctly and incorrectly detected iris pixels in a segmentation given by the boundary curves $P_i, L_i$ and the ground truth mask $G_i$ of image $i$. The mask, with dimension $m \times n$, is split into four groups (illustrated in Fig. 2) as follows: true positives ($tp_i$), the number of iris pixels which were correctly detected; false positives ($fp_i$), the number of non-iris pixels which were detected as iris pixels; false negative ($fn_i$), the number of iris pixels which were not detected; and true negative ($tn_i$), the number of correctly detected non-iris pixel.

We will use two sets of measures based on the $N$ images of size $m \times n$ in a given database. One is the set of measures suggested by the Noisy Iris Challenge Evaluation - Part

I (NICE.I) defined as:

$$E_1 := \frac{1}{N}\sum_{i=1}^{N}\frac{fp_i + fn_i}{mn}, \qquad (1)$$

$$E_2 := \frac{1}{2}\left(\frac{1}{N}\sum_{i=1}^{N}\frac{fp_i}{fp_i + tn_i}\right) + \frac{1}{2}\left(\frac{1}{N}\sum_{i=1}^{N}\frac{fn_i}{fn_i + tp_i}\right). \qquad (2)$$

The error rate $E_1$ refers of the rate of pixel disagreement between ground truth and segmentation noise masks (the classification error rate) and $E_2$ accounts for the disproportion between apriori probabilities (Type-I and Type-II error rate).

The other set of measures are well known from the field of information retrieval, e.g. [38]: The precision defined as

$$\mathscr{P} := \frac{1}{N}\sum_{i=1}^{N}\frac{tp_i}{tp_i + fp_i}, \qquad (3)$$

gives the percentage of retrieved iris pixels which are correct. The recall

$$\mathscr{R} := \frac{1}{N}\sum_{i=1}^{N}\frac{tp_i}{tp_i + fn_i}, \qquad (4)$$

gives the percentage of iris pixels in the ground truth which were correctly retrieved.

Since the target is to optimize both recall and precision these two scores are combined by the $\mathscr{F}$-measure, which is the harmonic mean of $\mathscr{P}$ and $\mathscr{R}$,

$$\mathscr{F} := \frac{1}{N}\sum_{i=1}^{N}\frac{2\mathscr{R}_i\mathscr{P}_i}{\mathscr{R}_i + \mathscr{P}_i}. \qquad (5)$$

The $\mathscr{F}$-measure, or to be more correct the $\mathscr{F}_1$-measure, is for our case equivalent to the Dice and Sørensen measures, which are also sometimes used for information retrieval.

## 3 Predictive Value of Segmentation Measures

As a first step we want to estimate how much the various measures differ. The result for the segmentation on the three databases is given in Table 1, more detailed information (mean and standard deviation) for the $\mathscr{F}$-measure, $\mathscr{P}$ and $\mathscr{R}$ can be found in [39]. From the table we can see that there is a difference in the measure for segmentation accuracy, although it seems only a small one. Since the measures are not interchangeable the measure which best represents the desired result of an segmentation algorithm should be chosen.

The second step then has to be to determine if, and which, measures might be able to predict the overall recognition accuracy of the iris tool chain based on segmentation. A large number of results for the different combinations of database, segmentation and feature extraction can be found in Table 2. The results are grouped by database and feature extraction method, per group the results are ordered by segmentation performance as given by the $\mathscr{F}$-measure. The performance of the system is given in terms of the receiver operating characteristic (ROC) curve. Since ROC plots would take up to much space we reduce the ROC to area under curve (AUC) and two operating points, the equal-error rate (EER) and 'OP 0.01' giving the false reject rate (FRR) for a fixed false acceptance rate (FAR) of 0.01%. Note that GST and Osiris on the *casia4i* database would be ordered differently for $E_{\{1,2\}}$, cf. Table 1, and are marked with an asterisk. There are a number of observations to be made:

- For prediction of the iris tool chain performance a monotonous relationship with the $\mathscr{F}$-measure is desired. That means a decrease in segmentation accuracy should result in a decrease in AUC and an increase in 'OP 0.01' and EER. The entries are ordered by the $\mathscr{F}$-measure with EER/AUC/'OP 0.01' rates which do not match the ordering printed with a red background, e.g., if for the EER a latter entry is lower than the marked entry which denotes a break in monotonicity. That is the red entries represent entries where the performance of the iris tool chain performance is not predicted correctly by the segmentation accuracy.
- By underlining we marked a pair of segmentation and feature extraction which result in inverse ordering for one of the ROC related entries. No matter which segmentation measure is used it would rank the segmentation in the same order. The results of the overall iris tool chain however are ranked inversely.

| DB | segmentation | [%] | $E_1$[%] | $E_2$[%] |
|---|---|---|---|---|
| *notredame* | Op. A | 100.00 | 0.00 | 0.00 |
| | Osiris | 92.24 | 1.31 | 7.48 |
| | GST | 89.22 | 1.88 | 11.17 |
| | WAHET | 87.63 | 2.36 | 14.08 |
| | IFPP | 85.35 | 2.79 | 16.79 |
| | CAHT | 80.51 | 3.67 | 20.57 |
| *casia4i* | Op. A | 100.00 | 0.00 | 0.00 |
| | Osiris | 92.74 | 5.22 | 9.72 |
| | GST | 90.97 | 4.56 | 8.80 |
| | CAHT | 89.27 | 6.29 | 12.17 |
| | WAHET | 89.13 | 6.06 | 11.60 |
| | IFPP | 84.98 | 7.71 | 14.51 |
| *iitd* | Op. A | 100.00 | 0.00 | 0.00 |
| | Osiris | 89.36 | 4.61 | 8.04 |
| | GST | 89.29 | 6.15 | 10.73 |
| | CAHT | 86.28 | 8.54 | 15.47 |
| | IFPP | 85.46 | 9.13 | 16.40 |
| | WAHET | 82.56 | 10.11 | 17.80 |

Table 1: Comparison of the various segmentation algorithms on the three test databases. The primary sorting is per $\mathscr{F}$-measure and disagreements in sorting of $\mathscr{F}$, $E_1$ and $E_2$ are marked with a red background.

This means that a segmentation measure can never predict the output of the whole iris biometric tool chain exactly. We can use correlation, based on rank since we can not assume linearity, to estimate the predictive performance of the $\mathscr{F}$-measure. This is done in Table 3 by using the Spearman rank order correlation (SROC). From the table we see that overall the prediction is dependant on the combination of database, feature and value-to-be-predicted. As an example take the lg feature extraction. The EER and AUC are decently predicted by the segmentation, but for 'OP 0.01' the prediction on *iitd* is not good. Similar behaviour can be seen for AUC with ko and EER with cr.
- It should also be noted that the different feature extraction methods show a non-uniform resilience to segmentation methods and databases:
    - On the *casia4i* database, the sift algorithm seems to handle all the segmentations, except IFPP, with a similar EERs. However, this behaviour seems to be non-uniform over different databases. Looking at the *iitd* database, sift seems to be able to handle the Osiris segmentation much better than the GST, CAHT and IFPP segmentations.
    - The dct algorithm has a huge problem with the GST segmentation on the *casia4i* database while it works well with GST on the *iitd* database.
    - Also, the magnitude of difference in the $\mathscr{F}$-measure does not correlate to the magnitude of the difference in the EER/AUC/'OP 0.01', i.e. the relation is non linear, Fig. 3 illustrates the EER over segmentation performance. For example take sift. On the *casia4i* database the difference in EER between Op. A and WAHET segmentation methods is small while the change in $\mathscr{F}$-measure is quite large. On the *iitd* database the difference in 'OP 0.01' between Op. A and CAHT is small while the change in $\mathscr{F}$-measure is large, similarly the change in $\mathscr{F}$-measure between CAHT and IFPP is small while the resulting change in 'OP 0.01' is large. For *notredame* the AUC between IFPP and CAHT is almost the same while the $\mathscr{F}$-measure would indicate a large change. This validates the use of a non-linear correlation coefficient in Table 3.
- The $\mathscr{F}$-measure seems to be better fit than $E_{\{1,2\}}$, for the given data. In Table 1 for the *casia4i* database the $E_{\{1,2\}}$ measures rate GST higher than Osiris, all other orderings are the same. In Table 2, where miss-matched orderings are marked with an asterisk, only the ko feature extraction method aligns with the ordering by $E_{\{1,2\}}$ for EER and 'OP 0.01'. All other features extraction methods agree with the $\mathscr{F}$-measure ordering of Osiris over GST. For AUC all tested feature extraction methods align with the ordering by the $\mathscr{F}$-measure.

So overall, while the segmentation measures do not agree, they also do not disagree in a substantial way. Also, the prediction of the overall performance of the iris biometrics tool chain is not possible. This said, the usage of segmentation measures for the analysis of the performance of segmentation algorithms can still be useful, as shown in [22, 40].

A note on the significance: Let us look at the EER, AUC and 'OP 0.01' of Operator A and Osiris with the qsw feature extraction. All three values are relatively close, so is this a significant difference?

Consider $H_0$: The Hamming distance of two iris codes are drawn from the same underlying (unknown) distribution. That means that the genuine scores based on the two segmentations methods are from a common distribution, i.e. there is no difference. This can be tested with the Kolmogorov Smirnov (KS) 2-sample statistics test [41]. The result gives a KS statistic of 0.134 for genuine which would reject $H_0$ with a p-value of $2 \times 10^{-70}$.

Furthermore, we could argue that while the distributions are different the result could still be the same, i.e. the overlap of the genuine and imposter distributions results in the same ROC. If that were the case values calculated from the ROC, like EER would show no statistically significant difference.

Let us consider the difference in EER operation points. If the ROC is the same then any difference between the two should be purely by chance. We can use the McNemar [42] test to find if this is true. That is, consider each comparison and classify it into a table:

|  |  | Operator A | |
|---|---|---|---|
|  |  | C | W |
| Osiris | C | A | B |
|  | W | C | D |

The fields are as follows: at the EER operation point A, is the number of correctly classified matches by both; B is the number of correctly classified by iris codes based on Osiris and incorrectly by those based on Operator A; C is the number of correctly classified iris codes based on Operator A and incorrectly by those based on Osiris; D is the number of wrongfully classified iris codes by both. So if the two ROC are basically the same then $H_0$: Distribution between B and C are random occurrences based on a binomial distribution with $p = q = 0.5$. The values for this specific case are $N = 3480841$, $A = 3430493$, $B = 22497$, $C = 25023$, $D = 2828$. Note that due to the high numbers we use the usual approximation by $X^2$ as well as Edward's correction. The result is $X^2 = 134.167192761$, for a p-value of 1% the required $X^2$ value is 6.64. This results in a rejection of $H_0$.

Even for these seemingly small differences in the results statistics show that the difference is significant.

## 4 Isolating Factors in Segmentation

Since the extent of deviation from a ground truth is apparently no indicator for the performance of the overall tool chain it would be interesting to know how the segmentation influences the iris tool chain. In this section we will modify the segmentation in a controlled way and give the results of the overall tool chain. This experimental evaluation results in a lot of data, which is very time consuming to produce, so we restrict the evaluation to a subset of the algorithms used before, specifically to WAHET and CAHT. We will also include the ground truth as a baseline. For feature extraction we will use the lg and qsw algorithms since they are consistently among the best of the tested algorithms in Table 2. For modification three factors, which follow directly from Daugman's rubbersheet transform, influence the normalisation of the iris, the correctness of the pupillary boundary, the correctness of the iris to sclera boundary and the normalisation center. It should be noted that in practice the normalisation center is a very stable parameter and is rarely miss-detected. Similarly, the pupillary boundary detection is very stable and, if any, only minor errors occur. The iris to sclera (limbic) boundary is the one which is most often miss-detected, mainly due to the obscuring influence of cilia and eyelids. It should also be noted that this only holds for near infrared

| segmen-tation | fea-ture | $\mathscr{F}$ [%] | EER [%] | AUC [%] | OP 0.01 [%] | segmen-tation | fea-ture | $\mathscr{F}$ [%] | EER [%] | AUC [%] | OP 0.01 [%] | segmen-tation | fea-ture | $\mathscr{F}$ [%] | EER [%] | AUC [%] | OP 0.01 [%] |
|---|---|---|---|---|---|---|---|---|---|---|---|---|---|---|---|---|---|
| | | *casia4i* | | | | | | *iitd* | | | | | | *notredame* | | | |
| Op. A | lg | 100.00 | 0.92 | 99.73 | 2.26 | Op. A | lg | 100.00 | 0.47 | 99.88 | 19.49 | Op. A | lg | 100.00 | 23.50 | 86.08 | 42.22 |
| Osiris | lg | *92.74 | 1.04 | 99.66 | 3.52 | Osiris | lg | 89.36 | 1.40 | 99.62 | 19.49 | Osiris | lg | 92.24 | 24.65 | 84.98 | 48.07 |
| GST | lg | *90.97 | 2.97 | 99.07 | 8.86 | GST | lg | 89.29 | 1.19 | 99.43 | 18.58 | GST | lg | 89.22 | 24.57 | 85.24 | 47.83 |
| CAHT | lg | 89.27 | 1.22 | 99.53 | 3.19 | CAHT | lg | 86.28 | 1.85 | 99.13 | 16.03 | WAHET | lg | 87.63 | 24.58 | 81.20 | 54.54 |
| WAHET | lg | 89.13 | 1.89 | 99.38 | 4.03 | IFPP | lg | 85.46 | 3.87 | 96.68 | 35.29 | IFPP | lg | 85.35 | 27.99 | 84.91 | 49.16 |
| IFPP | lg | 84.98 | 8.10 | 95.63 | 17.27 | WAHET | lg | 82.56 | 6.82 | 98.27 | 27.51 | CAHT | lg | 80.51 | 28.18 | 81.20 | 64.68 |
| Op. A | qsw | 100.00 | 0.61 | 99.85 | 1.24 | Op. A | qsw | 100.00 | 0.47 | 99.86 | 21.92 | Op. A | qsw | 100.00 | 22.95 | 86.48 | 36.67 |
| Osiris | qsw | *92.74 | 0.73 | 99.75 | 1.88 | Osiris | qsw | 89.36 | 1.21 | 99.68 | 19.86 | Osiris | qsw | 92.24 | 24.45 | 85.18 | 41.56 |
| GST | qsw | *90.97 | 2.40 | 99.38 | 6.80 | GST | qsw | 89.29 | 1.23 | 99.52 | 18.33 | GST | qsw | 89.22 | 23.77 | 85.89 | 42.92 |
| CAHT | qsw | 89.27 | 0.99 | 99.65 | 1.72 | CAHT | qsw | 86.28 | 1.72 | 99.15 | 14.24 | WAHET | qsw | 87.63 | 24.22 | 82.40 | 49.53 |
| WAHET | qsw | 89.13 | 1.72 | 99.43 | 2.91 | IFPP | qsw | 85.46 | 4.36 | 96.03 | 37.96 | IFPP | qsw | 85.35 | 27.32 | 85.46 | 43.87 |
| IFPP | qsw | 84.98 | 8.78 | 94.90 | 16.74 | WAHET | qsw | 82.56 | 7.43 | 97.99 | 29.90 | CAHT | qsw | 80.51 | 27.23 | 82.21 | 64.22 |
| Op. A | dct | 100.00 | 0.78 | 99.96 | 4.51 | Op. A | dct | 100.00 | 0.65 | 99.88 | 21.87 | Op. A | dct | 100.00 | 24.12 | 85.05 | 56.09 |
| Osiris | dct | *92.74 | 1.36 | 99.84 | 7.40 | Osiris | dct | 89.36 | 2.25 | 99.41 | 23.22 | Osiris | dct | 92.24 | 25.49 | 83.74 | 62.87 |
| GST | dct | *90.97 | 13.72 | 99.01 | 32.58 | GST | dct | 89.29 | 1.76 | 99.52 | 21.23 | GST | dct | 89.22 | 26.87 | 82.54 | 68.60 |
| CAHT | dct | 89.27 | 1.34 | 99.64 | 4.45 | CAHT | dct | 86.28 | 1.72 | 99.32 | 17.71 | WAHET | dct | 87.63 | 26.58 | 81.48 | 78.68 |
| WAHET | dct | 89.13 | 2.34 | 98.99 | 6.78 | IFPP | dct | 85.46 | 4.62 | 96.94 | 41.89 | IFPP | dct | 85.35 | 28.87 | 82.71 | 66.64 |
| IFPP | dct | 84.98 | 10.83 | 94.79 | 39.01 | WAHET | dct | 82.56 | 6.60 | 98.62 | 42.43 | CAHT | dct | 80.51 | 27.63 | 80.77 | 81.71 |
| Op. A | ko | 100.00 | 13.31 | 93.92 | 68.78 | Op. A | ko | 100.00 | 2.28 | 99.28 | 22.90 | Op. A | ko | 100.00 | 30.50 | 77.46 | 82.96 |
| Osiris | ko | *92.74 | 14.01 | 93.23 | 75.47 | Osiris | ko | 89.36 | 5.73 | 98.38 | 35.48 | Osiris | ko | 92.24 | 29.81 | 78.03 | 87.82 |
| GST | ko | *90.97 | 13.72 | 92.54 | 54.10 | GST | ko | 89.29 | 3.51 | 98.88 | 19.39 | GST | ko | 89.22 | 27.89 | 80.31 | 81.43 |
| CAHT | ko | 89.27 | 13.50 | 96.47 | 41.74 | CAHT | ko | 86.28 | 3.11 | 98.35 | 19.75 | WAHET | ko | 87.63 | 32.42 | 73.72 | 89.09 |
| WAHET | ko | 89.13 | 14.04 | 93.25 | 68.04 | IFPP | ko | 85.46 | 6.69 | 93.86 | 36.23 | IFPP | ko | 85.35 | 33.90 | 74.96 | 84.16 |
| IFPP | ko | 84.98 | 19.06 | 87.77 | 74.97 | WAHET | ko | 82.56 | 9.78 | 96.50 | 29.18 | CAHT | ko | 80.51 | 33.51 | 73.10 | 88.10 |
| Op. A | cr | 100.00 | 6.50 | 98.30 | 49.29 | Op. A | cr | 100.00 | 1.58 | 99.71 | 25.25 | Op. A | cr | 100.00 | 25.69 | 82.80 | 93.53 |
| Osiris | cr | *92.74 | 10.04 | 96.16 | 59.26 | Osiris | cr | 89.36 | 6.47 | 97.18 | 31.79 | Osiris | cr | 92.24 | 29.36 | 78.75 | 97.71 |
| GST | cr | *90.97 | 14.38 | 90.01 | 92.32 | GST | cr | 89.29 | 6.32 | 97.09 | 29.93 | GST | cr | 89.22 | 28.73 | 78.33 | 95.30 |
| CAHT | cr | 89.27 | 6.45 | 97.82 | 37.21 | CAHT | cr | 86.28 | 2.66 | 98.33 | 24.29 | WAHET | cr | 87.63 | 28.38 | 74.94 | 95.17 |
| WAHET | cr | 89.13 | 7.82 | 96.60 | 39.81 | IFPP | cr | 85.46 | 10.90 | 92.99 | 41.60 | IFPP | cr | 85.35 | 31.58 | 79.63 | 94.70 |
| IFPP | cr | 84.98 | 21.44 | 85.56 | 62.74 | WAHET | cr | 82.56 | 9.88 | 94.06 | 39.37 | CAHT | cr | 80.51 | 31.32 | 75.73 | 95.50 |
| Op. A | sift | 100.00 | 3.39 | 99.14 | 18.62 | Op. A | sift | 100.00 | 0.69 | 99.84 | 28.79 | Op. A | sift | 100.00 | 25.19 | 84.02 | 60.88 |
| Osiris | sift | *92.74 | 3.49 | 99.08 | 18.30 | Osiris | sift | 89.36 | 0.78 | 99.77 | 31.11 | Osiris | sift | 92.24 | 25.53 | 83.88 | 63.81 |
| GST | sift | *90.97 | 3.88 | 98.76 | 23.97 | GST | sift | 89.29 | 1.51 | 99.34 | 27.33 | GST | sift | 89.22 | 27.32 | 83.66 | 62.99 |
| CAHT | sift | 89.27 | 3.35 | 99.03 | 18.13 | CAHT | sift | 86.28 | 1.82 | 99.11 | 29.11 | WAHET | sift | 87.63 | 25.32 | 82.25 | 63.41 |
| WAHET | sift | 89.13 | 3.81 | 98.88 | 22.03 | IFPP | sift | 85.46 | 1.16 | 98.43 | 65.71 | IFPP | sift | 85.35 | 25.62 | 84.11 | 62.43 |
| IFPP | sift | 84.98 | 7.52 | 96.34 | 92.55 | WAHET | sift | 82.56 | 3.74 | 99.63 | 27.45 | CAHT | sift | 80.51 | 27.56 | 84.10 | 62.88 |

Table 2: Comparison of segmentation and feature extraction combinations methods based on segmentation accuracy ($\mathscr{F}$) and the overall performance in terms of EER, area under curve (AUC) and the FRR at FAR=0.01% (as 'OP 0.01'). The results are grouped by database and feature extraction method and ordered based on the $\mathscr{F}$-measure. The marks and their meaning is described and discussed on page 4.

| DB | feature | EER | AUC | OP 0.01 |
|---|---|---|---|---|
| casia4i | lg | 0.829 | 0.829 | 0.714 |
| | qsw | 0.829 | 0.829 | 0.714 |
| | dct | 0.543 | 0.943 | 0.371 |
| | ko | 0.771 | 0.371 | 0.143 |
| | cr | 0.371 | 0.543 | 0.029 |
| | sift | 0.543 | 0.829 | 0.486 |
| iitd | lg | 0.943 | 0.943 | 0.386 |
| | qsw | 1.000 | 0.943 | 0.371 |
| | dct | 0.771 | 0.886 | 0.486 |
| | ko | 0.771 | 0.886 | 0.257 |
| | cr | 0.714 | 0.771 | 0.543 |
| | sift | 0.829 | 0.657 | 0.029 |
| notredame | lg | 0.829 | 0.843 | 0.886 |
| | qsw | 0.771 | 0.714 | 0.943 |
| | dct | 0.886 | 0.829 | 0.829 |
| | ko | 0.714 | 0.714 | 0.486 |
| | cr | 0.714 | 0.543 | 0.200 |
| | sift | 0.714 | 0.371 | 0.029 |

Table 3: Spearman rank order correlation (SROC) between the $\mathscr{F}$-measure and EER, AUC and 'OP 0.01'.

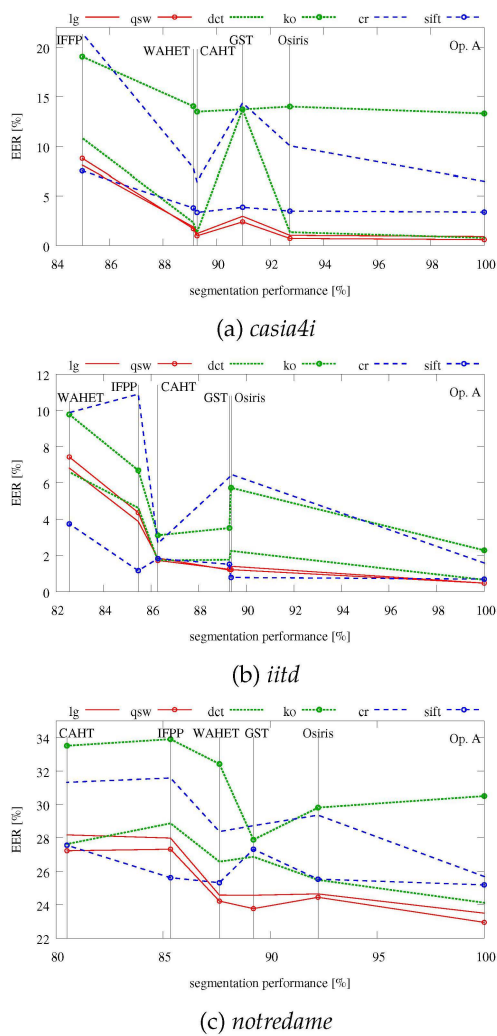

(a) *casia4i*

(b) *iitd*

(c) *notredame*

Figure 3: The equal error rate plotted over the segmentation performance ($\mathscr{F}$-measure) to highlight the nonlinear relation between segmentation performance and EER.

image types, of which all three databases consist, and does not hold for visible light iris images.

In our experiments we will use the masks as detected by the given algorithm and artificially scale the boundaries in such a way that the area is a percentage of the original iris. That is, the limbic boundary $L_{0.5}$ is a scaling of the limbic boundary such that the area of $L_{0.5}$ is equal to 0.5 times the area of the original $L_1$ boundary. The translation is based on the (unscaled) limbic boundary radius. We translate only along the x-axis and the $C_{0.5}$ is shifted along the x-axis for half the radius of the limbic boundary, i.e. $C_1$, would result in the normalisation center begin located on the limbic boundary. The modifications are illustrated in Fig. 4.

## 4.1 Scaling the Limbic Boundary

The iris texture at the limbic boundary is frequently obfuscated by eyelids and cilia. Furthermore, from [43–46] it is known that reliable iris texture information tends to be found at the pupillary boundary. When scaling the limbic boundary the expectations would be that scaling down should only show a minor influence in the results since mostly unusable iris texture is removed. On the other hand, scaling up would introduce more noise in the resulting iris texture and a compression of the more informative texture parts near the pupillary boundary. Figure 5 gives the results of the limbic boundary scaling tests. It is interesting to see the influence of scaling on the different databases. For the *casia4i* database the results follow the expectations very well. For the *iitd* database the case is slightly different, an upscaling does not influence the results much, while downscaling only shows a minor negative effect. Similar to the *iitd* case the *notredame* database shows little influence on scaling up but a more pronounced influence when scaling down. While it is unclear why exactly these different types of influences happen, the overall behavior confirms the sentiments of Section 3 that the performance of the tool chain exhibits a non-uniform response to the segmentation and database.

## 4.2 Scaling the Pupillary Boundary

When scaling the pupillary boundary the expectation would be: Scaling up will cut of the high informative texture information at the pupillary boundary and should reduce the overall performances; Scaling down will leave the texture information intact but compresses it and should thus influence the result to a lesser degree. The result of the experiment is given in Figure 6. The *casia4i* again conforms to expectations and shows that both scaling directions influence the results negatively but scaling down does so to a slightly lesser degree. And again *iitd* and *notredame* behave in a counter-intuitive way. What is especially interesting is that cutting off the high information texture area in the case of the *iitd* database seems to actually improve the results slightly. Again, the reason for this behavior is unclear, however, the difference in behavior on the different databases makes clear that an algorithm, or the whole iris biometric tool chain, should never be tested on a single database, since a generalization of the findings is clearly not possible.

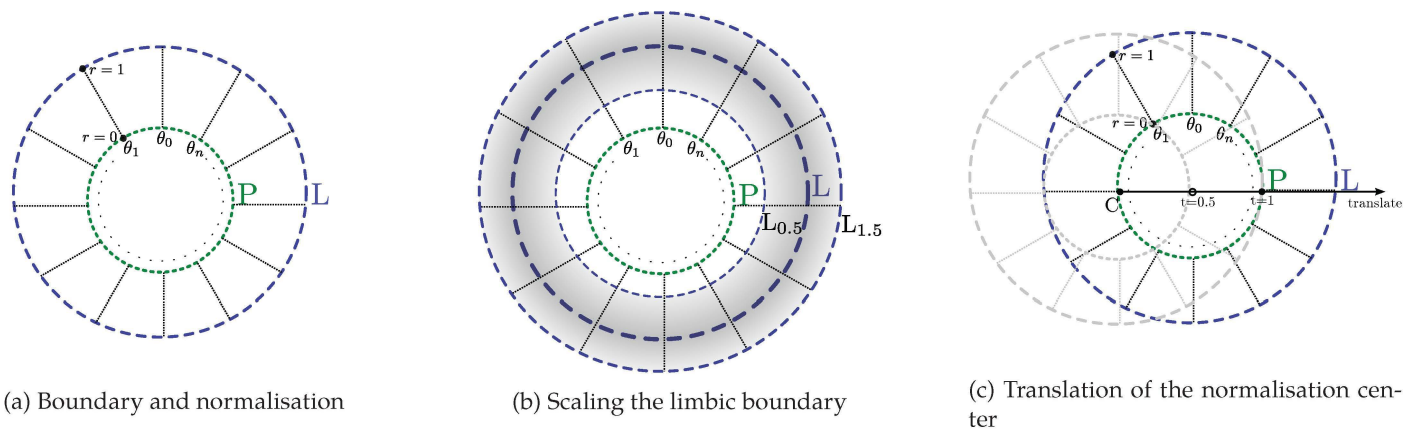

(a) Boundary and normalisation  (b) Scaling the limbic boundary  (c) Translation of the normalisation center

Figure 4: Schematic overview over iris boundary scaling and center point translation.

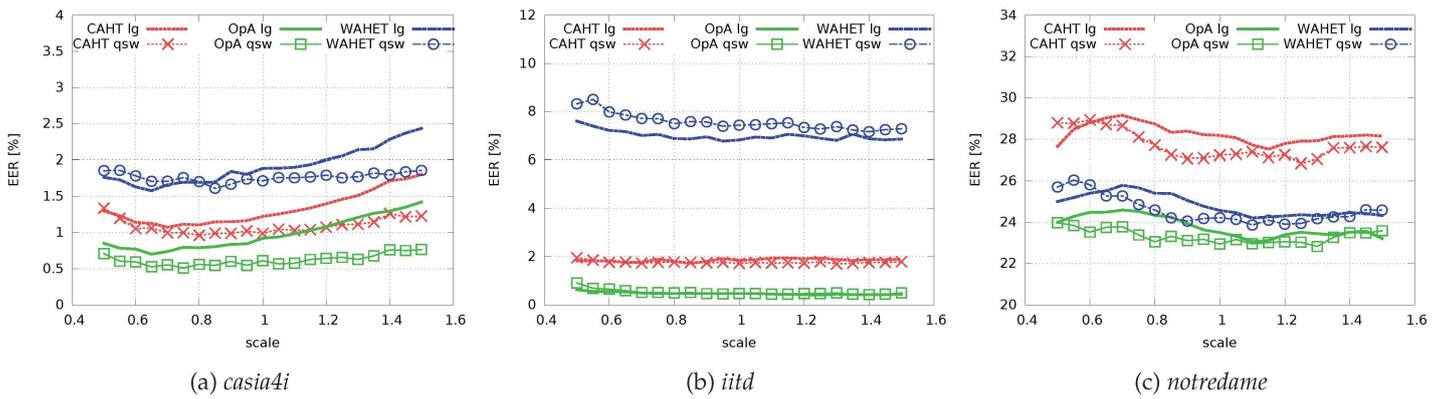

(a) *casia4i*  (b) *iitd*  (c) *notredame*

Figure 5: Scaling of the limbic boundary and the effects on the EER.

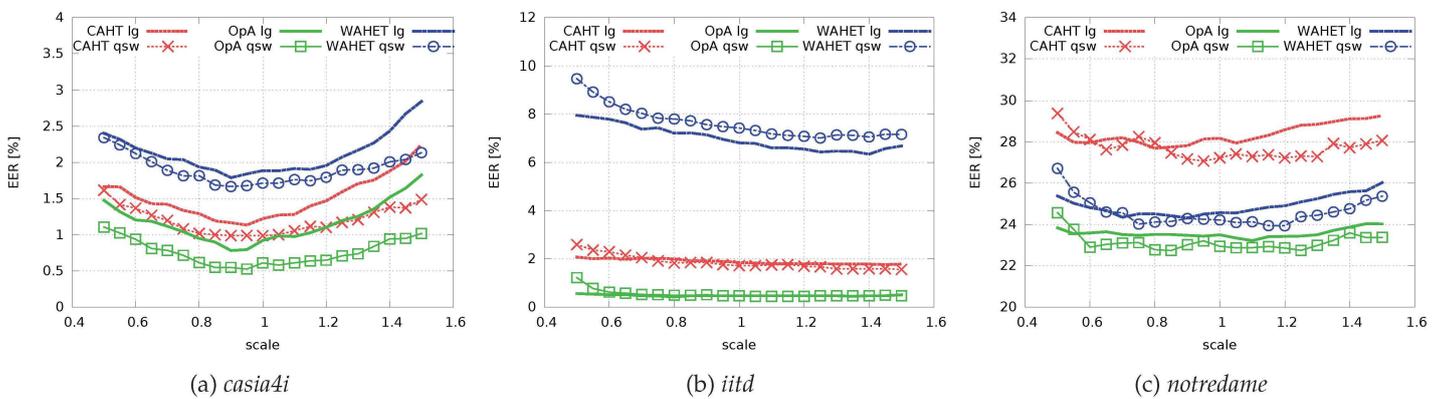

(a) *casia4i*  (b) *iitd*  (c) *notredame*

Figure 6: Scaling of the pupillary boundary and the effects on the EER.

### 4.3 Iris Center Detection

Translating the iris center influences the normalisation and consequently disallows for rotational alignment of texture information by horizontal shifting of the iris textures or iris codes. As such, the more the normalisation center is offset from the original iris center the worse the performance of the iris tool chain should become. The experimental results are given in Figure 7. On the *casia4i* database all combinations of tested segmentation and feature extraction algorithms behave exactly as expected with very little difference between combinations. For this test the *iitd* and *notredame* databases also exhibit a similar behavior, although less pronounced. For the *iitd* database an interesting effect is the difference between the WAHET and CAHT/Op. A behavior. While the WAHET clearly is strongly affected by the translation of the normalisation center, the CAHT and Op. A segmentations show an almost negligible effect for small offsets. This is especially interesting when compared to the almost uniform behavior of all segmentation methods on the *casia4i* database. Another noteworthy effect shows itself in the *notredame* database, where a slight offset of the center seems to improve performance. An improvement in a one-sided offset could indicate a systemic miss-detection of the normalisation center (CAHT). But the behavior of Op. A and WAHET is symmetric so such an explanation is not possible. Again the behavior of a change in segmentation results in an non-uniform response in performance. Also, the translation of the normalisation center clearly has the highest impact of all three tested modifications.

### 4.4 Segmentation Consistency versus Correctness

From the previous sections we can see that minor changes in segmentation accuracy do not influence the results overly much, with the exception of the normalisation center. However, the modifications were global, i.e., the same modification was performed on the whole database. In Wild *et al.* [40] mask based fusion was done in order to improve segmentation and the results were analysed. The authors reported that improvement took place when the segmentation errors were corrected. They specifically mentioned that the mask fusion, when improving overall results, notably improved on the low quality results while leaving high quality results virtually the same. This raises the question about consistency versus accuracy. The experiments up to now only dealt with segmentation accuracy and left consistency out of the picture since all segmentation results were modified in exactly the same way.

Consistency in this context means that the iris codes, and textures, are based on a correct segmentation of an iris image. Inconsistencies then are different segmentations, e.g., one image is correctly segmented, the whole iris is detected, the other one is incorrectly segmented, for example by miss-detection of the collarette for the limbic boundary.

Let us briefly consider the outliers, as specified in [22]. A z-score is calculated for every image $I$ in the database $\mathbf{D}$:

$$z(I, \mathbf{D}) = \frac{\mathscr{F}(I) - \mu(\mathscr{F}_{i \in \mathbf{D}}(i))}{\sigma(\mathscr{F}_{i \in \mathbf{D}}(i))},$$

with an image counting as outlier if $|z| > 3$. Note that even if we take the number of outliers, we still can not predict the overall performance, following the results from Section 3. However, usually optimization for segmentation is done based on accordance with a ground truth and accuracy is reported. It is however not quite known how important consistency is. Meaning what is the influence of outliers during segmentation and how robust are feature extraction algorithms to inconsistencies.

Since the amount of experiments required is high we will restrict experimentation to the *casia4i* database. From prior sections we know the *casia4i* shows the best performance, so the effects shown in this section should be expected to be more pronounced on the *iitd* and *notredame* databases.

Figures 8 and 9 compare a difference in scaling for the limbic and pupillary boundary respectively. The compared iris codes are results from two differently segmented iris images. Two steps of scaling difference are compared to the zero-difference scaling version from prior sections. The scaling is always towards the unscaled version, so at scale ±0.1 the result given at the 1.1 scale is actually the results of comparing the 1.1 scaled segmentation to the unscaled version (factor 1), similarly the point at scale 0.5 is the comparison of the scale 0.5 to scale 0.6.

From the figures we can see that a small change in scale, i.e., ±0.05 scale difference exhibits almost no impact on the performance. For the larger scale difference the impact is more pronounced. From Fig. 8c we can see that performance impact is low for the upscaled versions (scales > 1) while the performance impact for downscaled comparisons (scales < 1) is quite obvious. That is, a comparison between segmentations with small scale differences (Fig. 8b) behaves almost the same as comparisons without a scale difference (Fig. 8a) but larger scale differences impact the performance. Since a different amount of iris textures is stretched to a fixed size—the line segment between the pupillary and the limbic boundary changes in length but is scaled to the same fixed height iris texture—during the normalisation, the remaining information is distributed differently and thus impacts the resulting performance. The same effect should then be observable for the upscale version of the pupillary boundary. If anything the effect should be more pronounced since not only the amount of texture is reduced, but the highly informative texture information from the vicinity of the pupillary boundary is also removed. Figure 9c does not show this at all, rather the comparisons for the two scale differences, ±0.05 and ±0.1, behave very similarly. If we do a further doubling in scale difference to ±0.2 the same errors are apparent but to much higher degree (note that the range of the y-axis for Fig. 8d is different from the rest). It especially interesting that the downscaled versions of the limbic boundary results in a strong change in the iris code, and consequently the recognition rate. This is a strong indicator that possible obfuscations can result in a strong difference in features.

To get a better understanding of the influence of segmentation consistency we will look at two specific examples: User 1229 right eye segmented with the CAHT algorithm and user 1022 left eye segmented with the WAHET algorithm.

Figure 10 show the results for the right eye of user 1229, with relevant segmentation results. Only two images from this users iris are miss-segmented, Figure 10a shows a correctly segmented image. Figure 10b shows a slight miss segmentation, similar to the above experiments, and Fig. 10c shows a very strong segmentation error. The distribution plot in Fig. 10d shows the overall response/ frequency distribution of the iris tool chain for imposter and genuine comparisons, with EER and individual responses for comparisons of the iris image from user 1229 marked. Unsurprisingly comparisons with the miss-detected iris image (ID 3) shows a strong increase in response pushing the results clearly towards rejection. At the same time correctly segmented images stay clearly inside the accept range. However, even the slight miss-detection (ID 6), responses are labeled by comparison and shown as dash/dotted lines, shows a clear increase in response in comparison to the correctly segmented iris images.

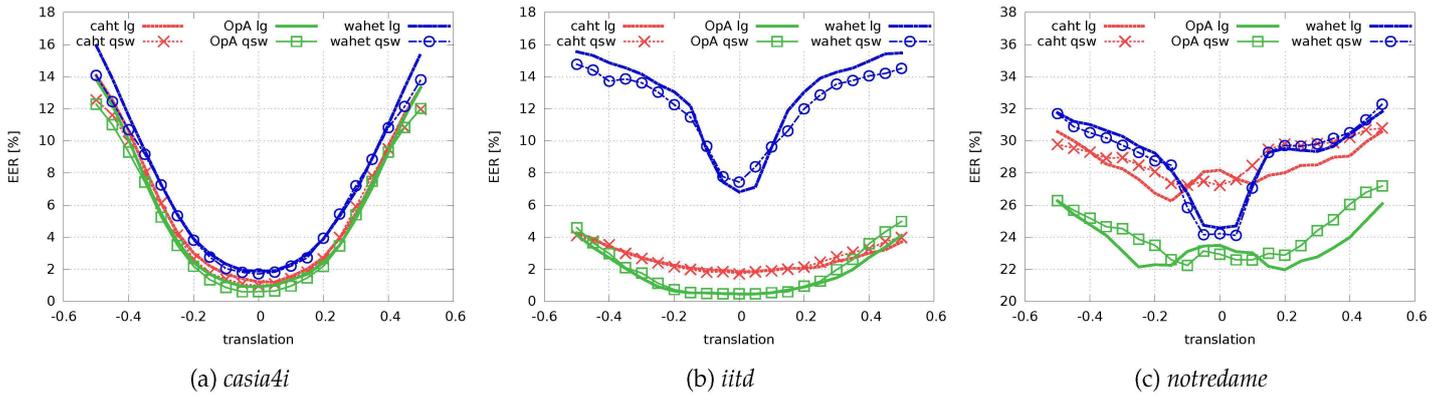

(a) *casia4i*  (b) *iitd*  (c) *notredame*

Figure 7: Translation of the iris center and the effects on the EER.

It seems that small inconsistencies only introduce a small error, but an error nonetheless. This suggests that the response is strongly tied to the extent of miss-detection, and an indicator that consistency is important. It seems, that small inconsistencies only result in a small change in response, explaining the results of figures 8 and 9.

In order to take a closer look at the effects of accuracy versus consistency let us take a look at Figure 11, user 1022, left eye segmented with WAHET. The database contains five iris images of this users left eye, which can be split into two groups: ID 1 and ID 3, Fig. 11a, which only suffer a slight miss-detection, and a group consisting of IDs 2,4, and 5, Fig. 11b which miss-detected the collarette for the limbic boundary. The interesting part of this case is that the two groups are internally consistent. The results of the frequency/response plot with marked comparison scores is given in Fig. 11c. The intra-group comparisons show a good response, and the results from full and reduced iris-textures are very comparable. Inter-group responses however show a clear change in response. This clearly shows that accuracy is of high importance, and reaffirms the findings in [40]. This behavior also explains the relatively low impact of the scaling of inner and outer iris boundaries in Figures 5 and 6. Overall, consistency seems to be more important than accuracy. In practice, an algorithm should not only aim for a high average segmentation accuracy but also for a low number of outliers. It is also conceivable to target the collarette instead of the limbic boundary as a segmentation target. Figure 5 shows that a reduction of the iris-texture to this area is unproblematic, and the collarette is less obfuscated by cilia and eyelids and might make for an easier segmentation target. This result is similar to the findings of Islam et al. [44], who found that, for their use of the Daubechies wavelet transform, the pupillary to collarette region has the most discriminating power.

It should be noted that the detection error when using the collarette has to be smaller. This follows directly from Fig. 8 where it can be seen that errors at the limbic boundary and upscaling, $L_x, x > 1$, hardly affect the results but for downscaled version, $L_x, x < 1$, the difference in scale, i.e., detection error, has a larger influence. From the figure we see that about ±0.05 is a save difference, at ±0.1% a clear deterioration of the EER is obvious. Note that this is in terms of area relative to the limbic boundary. Assuming that the collarette lies at $r_C = 0.8 \times r_L$ of the limbic radius $r_L$, this would require the $\hat{r}_C$ found by collarette segmentation to be within ±2.5% of the real radius $r_C$.

## 5 Conclusion and Future Work

We have shown that for ascertaining segmentation accuracy the choice of measure does not really matter since the differences are small, that said, the $\mathscr{F}$-measure seems to be the better choice. Furthermore, the segmentation accuracy is not a reliable predictor of overall iris tool chain performance, and the combination of database, segmentation and feature extraction behave non-uniformly, i.e., a feature extraction method can produce better overall results with a worse, based on segmentation accuracy, segmentation algorithm. This means that the choice of segmentation and feature extraction should not be made in isolation.

Regarding segmentation accuracy we found that it is not required to extract the whole iris image as long as the extracted region is consistent, i.e., the same region is extracted for both matched templates. This is valid under the assumption that the normalisation center is stable and the center point is correctly identified.

It should be noted that this is a result which does not conform with prior published results. Specifically, [47] did an analysis "independent of the choice of the segmentation algorithm" based on three feature extraction methods which looked into the influence of scaling and translation of the limbic and pupillary boundary by up to 5 pixels. Their find-

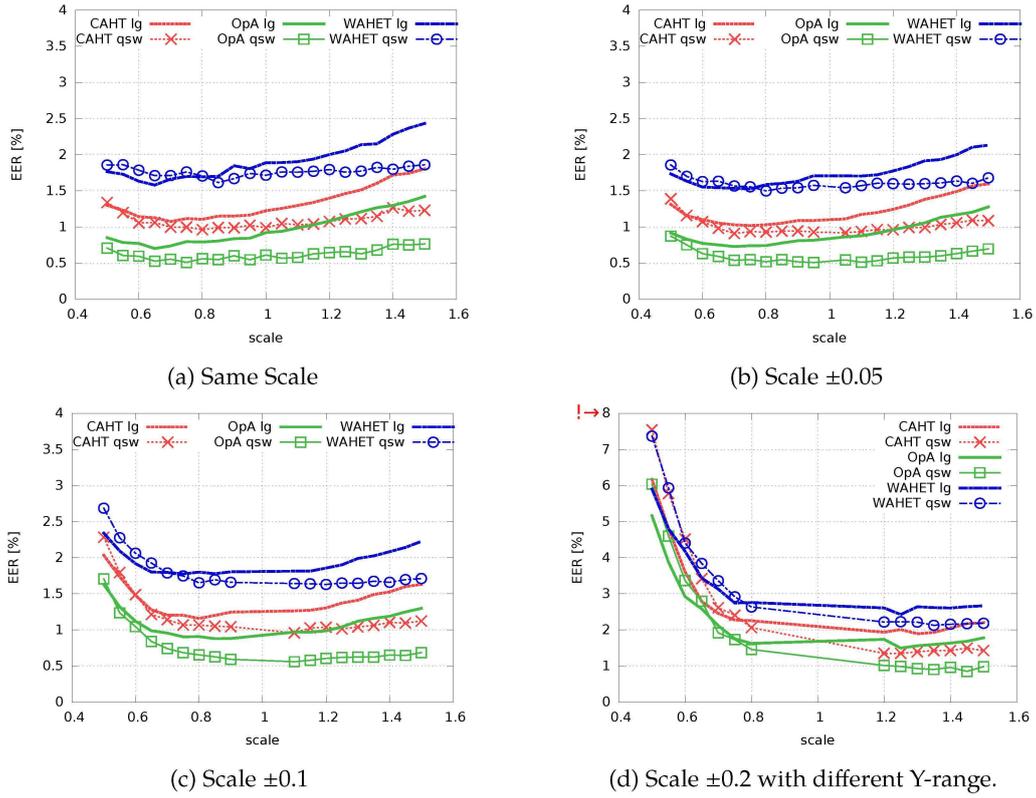

Figure 8: Scaling of the limbic boundary and comparing between scales on the *casia4i* database.

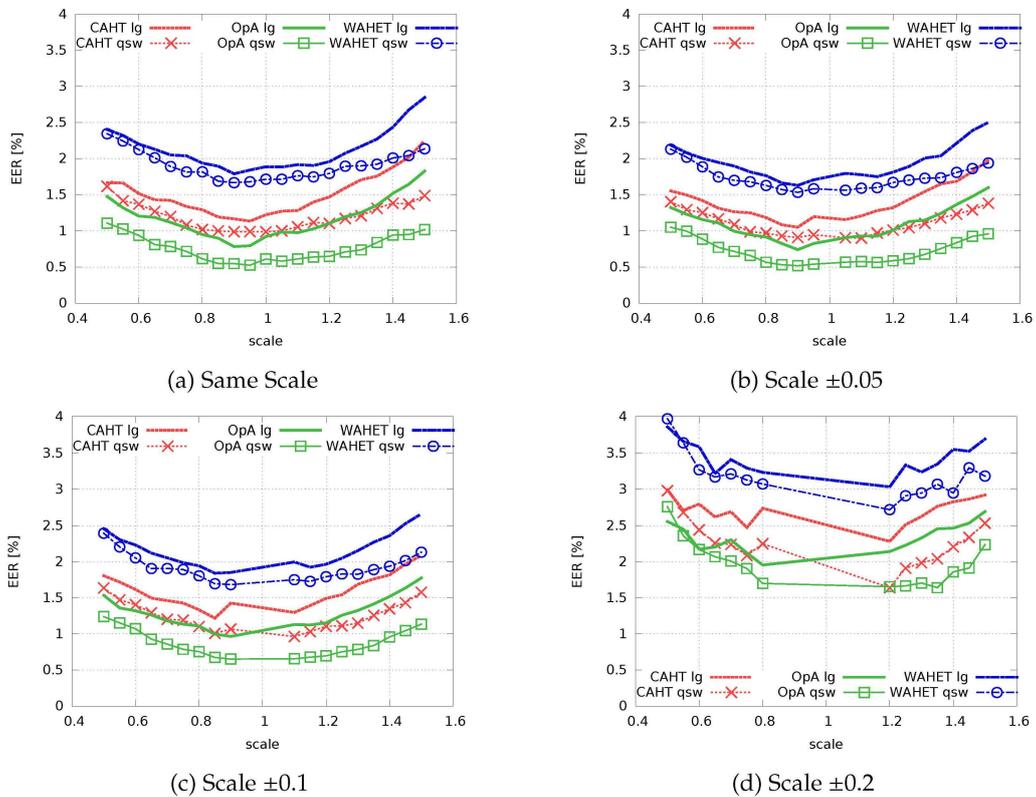

Figure 9: Scaling of the pupillary boundary and comparing between scales on the *casia4i* database.

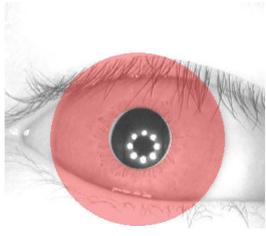
(a) Correct outer iris (ID 1).

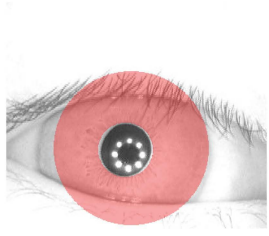
(b) Minor outer iris detection error (ID 6).

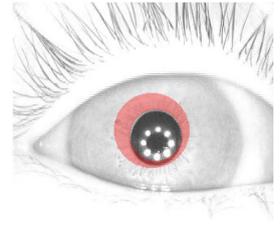
(c) Faulty detection (ID 3).

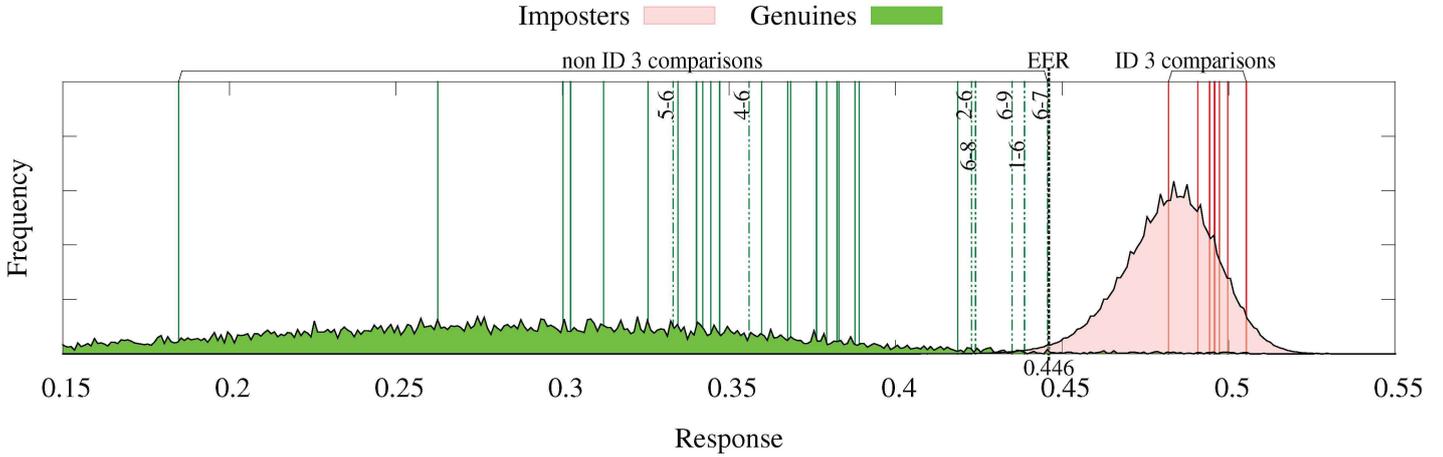
(d) Distribution plot and comparisons between correctly segmented and incorrectly segmented iris images.

Figure 10: Comparison response differentiated between intra and inter group comparison for the ID sets {1,3} and {2,4,5} for the CAHT segmentation of eye S1229R and the lg feature extraction.

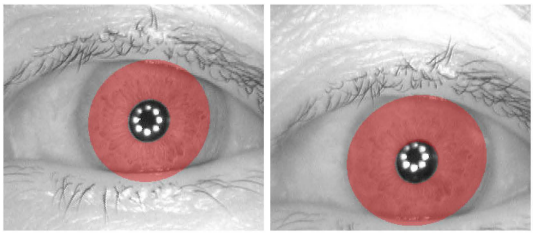
(a) Almost correct outer iris (ID 1 and 3).

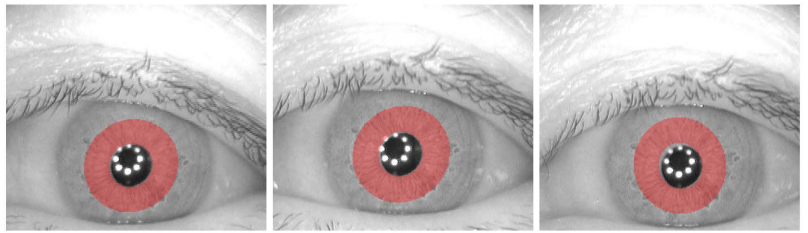
(b) Collarette miss-detection (IDs 2, 4 and 5).

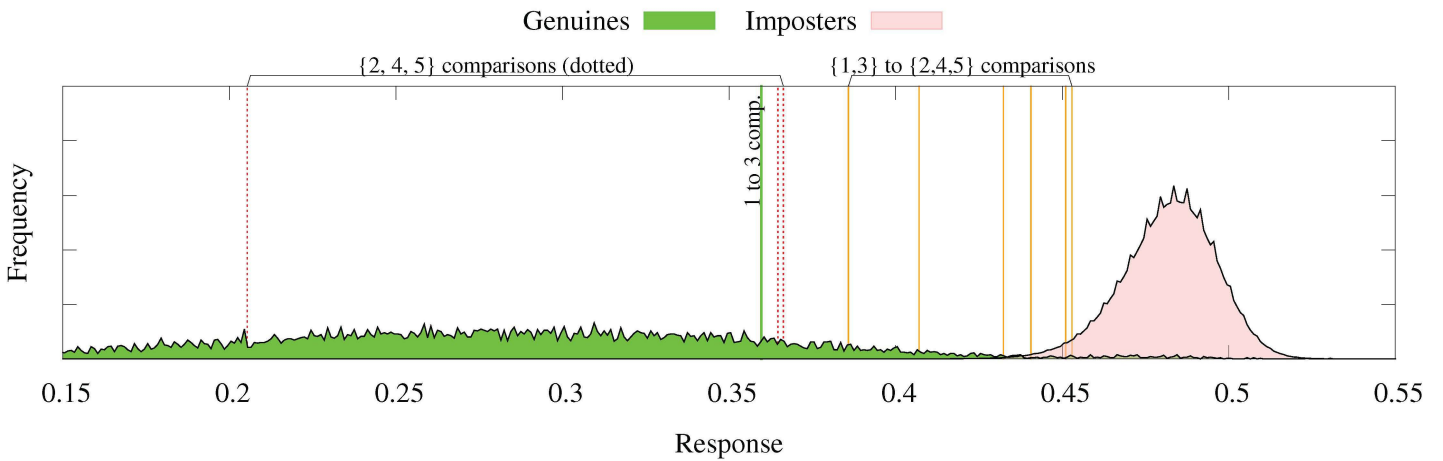
(c) Distribution plot and comparisons for correctly and incorrectly segmented iris images.

Figure 11: Comparison response differentiated between intra and inter groups for the ID sets {1,3} and {2,4,5} for the WAHET segmentation of eye S1022L and the lg feature extraction.

ing was "the increment in the error rates is directly proportional to the amplitude of the segmentation inaccuracies". It is difficult to ascertain why the results differ, especially since the paper used 3 feature extraction methods but only gave a single set of results, which can only be assumed to be an aggregate of the three methods. Given that they find an almost linear relationship we would assume that the comparison was between correct and erroneous segmentation. In this case the findings are somewhat similar, although they report a far higher impact on the error rates. Since the segmentation and feature extraction are not independent the differences may also be due to the segmentation they used.

Another interesting result from literature is [10] where three segmentation algorithms, and lg feature extraction, were compared. Interestingly they used classifiers for recognition, specifically KNN, MLP, Bayes based, rule based and SVM classifiers. They found a correlation between segmentation error and performance of the classifiers. As such it should be noted that our findings may also be dependant on the classifier.

## 5.1 Future Work

The experimental results in this paper have clearly shown that the combination of database, segmentation and feature extraction result in a non-uniform behavior. This effect is not clearly understood and it is of interest why, and how, the different combinations produce such, sometimes widely different, results.

A further effect which can be of interest is the influence of the rotation on the comparison. The problem is that the databases are not rotationally aligned. It should be possible to align the databases based on optimal alignment of iris codes. When such an alignment has been done the effect of the rotation can be studied.

## Acknowledgements

This work was partially supported by the Austrian Science Fund, project no. P27776.## References

[1] H. Proença and L. Alexandre, "Toward covert iris biometric recognition: Experimental results from the NICE contests," *IEEE Transactions on Information Forensics and Security*, vol. 7, no. 2, 2012. DOI: 10.1109/TIFS.2011.2177659 (cit. on p. 2).

[2] Z. He, T. Tan, Z. Sun, and X. Qiu, "Toward accurate and fast iris segmentation for iris biometrics," *Pattern Analysis and Machine Intelligence, IEEE Transactions on*, vol. 31, no. 9, pp. 1670–1684, Sep. 2009. DOI: 10.1109/TPAMI.2008.183 (cit. on p. 2).

[3] L. Ma, T. Tan, Y. Wang, and D. Zhang, "Efficient iris recognition by characterizing key local variations," *IEEE Transactions on Image Processing*, vol. 13, pp. 739–750, 2004 (cit. on pp. 2, 3).

[4] E. Tabassi, P. Grother, and W. Salamon, "Irex ii – iris quality calibration and evaluation," National Institute of Standards and Technology (NIST), Tech. Rep. NIST Interagency Report 7296, 2011 (cit. on p. 2).

[5] D. Benini and et al., "ISO/IEC 29794-6 Biometric Sample Quality - part 6: Iris image data," International Organization for Standardization, Tech. Rep. JTC1/SC37/Working Group 3, 2012 (cit. on p. 2).

[6] N. Zuo and N. A. Schmid, "An automatic algorithm for evaluation of precision of iris segmentation," in *2nd IEEE International Conference on Biometrics: Theory, Applications and Systems (BTAS 2008)*, Oct. 2008, pp. 1–6 (cit. on p. 2).

[7] F. Alonso-Fernandez and J. Bigun, "Quality factors affecting iris segmentation and matching," in *Biometrics (ICB), 2013 International Conference on*, 2013, pp. 1–6. DOI: 10.1109/ICB.2013.6613016 (cit. on p. 2).

[8] N. Kalka, J. Zuo, N. Schmid, and B. Cukic, "Estimating and fusing quality factors for iris biometric images," *Systems, Man and Cybernetics, Part A: Systems and Humans, IEEE Transactions on*, vol. 40, no. 3, pp. 509–524, May 2010. DOI: 10.1109/TSMCA.2010.2041658 (cit. on p. 2).

[9] M. Fairhurst and M. Erbilek, "Analysis of physical ageing effects in iris biometrics," *IET Computer Vision*, vol. 5, no. 6, pp. 358–366, 2011, ISSN: 1751-9632. DOI: 10.1049/iet-cvi.2010.0165 (cit. on p. 2).

[10] M. Erbilek, M. C. D. C. Abreu, and M. Fairhurst, "Optimal configuration strategies for iris recognition processing," in *IET Conference on Image Processing (IPR 2012)*, 2012, 6 p. (Cit. on pp. 2, 13).

[11] J. Daugman, "How iris recognition works," *IEEE Transactions on Circuits and Systems for Video Technology*, vol. 14, no. 1, pp. 21–30, 2004 (cit. on p. 2).

[12] R. P. Wildes, "Iris recognition: an emerging biometric technology," in *Proceedings of the IEEE*, vol. 85, 1997, pp. 1348–1363 (cit. on p. 2).

[13] K. Bowyer, K. Hollingsworth, and P. Flinn, "Image understanding for iris biometrics: A survey," *Computer Vision and Image Understanding*, vol. 110, no. 2, pp. 281–307, 2008 (cit. on p. 2).

[14] J. Daugman, "New methods in iris recognition," *Systems, Man, and Cybernetics, Part B: Cybernetics, IEEE Transactions on*, vol. 37, no. 5, pp. 1167–1175, Oct. 2007. DOI: 10.1109/TSMCB.2007.903540 (cit. on p. 2).

[15] D. S. Jeong, J. W. Hwang, B. J. Kang, K. R. Park, C. S. Won, D.-K. Park, and J. Kim, "A new iris segmentation method for non-ideal iris images," *Image and Vision Computing*, vol. 28, no. 2, pp. 254–260, 2010. DOI: http://dx.doi.org/10.1016/j.imavis.2009.04.001. [Online]. Available: http://www.sciencedirect.com/science/article/pii/S0262885609000638 (cit. on p. 2).

[16] S. Pundlik, D. Woodard, and S. Birchfield, "Iris segmentation in non-ideal images using graph cuts," *Image and Vision Computing*, vol. 28, no. 12, pp. 1671–1681, 2010. DOI: http://dx.doi.org/10.1016/j.imavis.2010.05.004. [Online]. Available: http://www.sciencedirect.com/science/article/pii/S026288561000079X (cit. on p. 2).

[17] H. Proenca, "Iris recognition: On the segmentation of degraded images acquired in the visible wavelength," *Pattern Analysis and Machine Intelligence, IEEE Transactions on*, vol. 32, no. 8, pp. 1502–1516, Aug. 2010. DOI: 10.1109/TPAMI.2009.140 (cit. on p. 2).

[18] K. Bowyer, K. Hollingsworth, and P. Flynn, "A survey of iris biometrics research: 2008–2010," in *Handbook of Iris Recognition*, ser. Advances in Computer Vision and Pattern Recognition, Springer London, 2013, pp. 15–54. DOI: 10.1007/978-1-4471-4402-1_2. [Online]. Available: http://dx.doi.org/10.1007/978-1-4471-4402-1_2 (cit. on p. 2).
13